\pdfoutput=1

\documentclass[11pt]{article}

\usepackage[final]{acl}

\usepackage{times}
\usepackage{latexsym}

\usepackage[T1]{fontenc}

\usepackage[utf8]{inputenc}

\usepackage{microtype}

\usepackage{inconsolata}

\usepackage{graphicx}

\usepackage{algorithm}
\usepackage{algpseudocode}  
\usepackage{amsmath}

\usepackage{latexsym}
\usepackage{amssymb}
\usepackage{amsmath}
\usepackage{amsthm}
\usepackage{booktabs}
\usepackage{enumitem}
\usepackage{graphicx}
\usepackage{color}

\usepackage{multirow} 
\usepackage{adjustbox} 
\usepackage{xspace}

\title{Not All Parameters Are Created Equal: Smart Isolation Boosts Fine-Tuning Performance}

\author{
  Yao Wang\textsuperscript{1}
  ,
  Di Liang\textsuperscript{2}
  ,
  Minlong Peng\textsuperscript{3}
  \\
  \textsuperscript{1}University of New South Wales\\
  \textsuperscript{2}ByteDance Inc.\\
  \textsuperscript{3}Fudan University\\
  \texttt{\{yao.wang11@student.unsw.edu.au, liangd17@fudan.edu.cn\}}
}

\begin{document}
\maketitle
\footnotetext[1]{Corresponding author: Minlong Peng.}
\footnotetext[2]{This work was done while Yao Wang was interning at ByteDance  under Di Liang's supervision.}
\begin{abstract}
Supervised fine-tuning (SFT) is a pivotal approach to adapting large language models (LLMs) for downstream tasks; however, performance often suffers from the ``seesaw phenomenon'', where indiscriminate parameter updates yield progress on certain tasks at the expense of others. To address this challenge, we propose a novel \emph{Core Parameter Isolation Fine-Tuning} (CPI-FT) framework. Specifically, we first independently fine-tune the LLM on each task to identify its core parameter regions by quantifying parameter update magnitudes. Tasks with similar core regions are then grouped based on region overlap, forming clusters for joint modeling. We further introduce a parameter fusion technique: for each task, core parameters from its individually fine-tuned model are directly transplanted into a unified backbone, while non-core parameters from different tasks are smoothly integrated via Spherical Linear Interpolation (SLERP), mitigating destructive interference. A lightweight, pipelined SFT training phase using mixed-task data is subsequently employed, while freezing core regions from prior tasks to prevent catastrophic forgetting. Extensive experiments on multiple public benchmarks demonstrate that our approach significantly alleviates task interference and forgetting, consistently outperforming vanilla multi-task and multi-stage fine-tuning baselines.
\end{abstract}

\section{Introduction}
\label{sec:introduction}



Large Language Models (LLMs) \cite{brown2020language, chowdhery2023palm, raffel2020exploring, touvron2023llama} have demonstrated remarkable generalization across diverse natural language tasks, achieving impressive success on benchmarks spanning reasoning, dialogue, instruction following, and more. SFT \cite{chung2024scaling, ouyang2022training, sanh2021multitask} remains a crucial methodology for tailoring these models to specific applications, aligning them with human instructions, and imbuing domain-specific expertise by optimizing on datasets of task-relevant examples.


Supervised fine-tuning (SFT) faces significant challenges in multi-task and multi-domain scenarios. When applied to heterogeneous datasets, such as mathematical reasoning, creative writing, coding, and factual question answering, conflicting optimization objectives among tasks often lead to the "seesaw effect" \cite{yu2020gradient}, where performance improvements on one task degrade others. This issue hinders the development of robust, broadly capable large language models (LLMs). Existing approaches, including joint multi-task fine-tuning, naive parameter sharing, and staged curricula \cite{ouyang2022training, caruana1997multitask, wei2021finetuned}, generally assume uniform parameter importance across tasks, updating all parameters indiscriminately. While multi-stage training alleviates direct gradient conflicts through sequential task structuring, it remains a coarse-grained isolation strategy that exacerbates catastrophic forgetting \cite{mccloskey1989catastrophic, kirkpatrick2017overcoming}, further eroding the model’s ability to generalize across diverse tasks.


We hypothesize that the root cause of these challenges lies in the phenomenon of \textit{parameter heterogeneity}: distinct capabilities of large language models (LLMs) rely on specific and potentially overlapping subsets of parameters, with certain clusters disproportionately contributing to particular tasks. Uniform updates across the entire parameter space fail to account for the specialized roles of these localized parameter subsets, thereby fostering destructive interference among competing tasks \cite{chen2018gradnorm}. Mitigating such interference necessitates a paradigm shift from heuristic approaches or task-level isolation to a principled framework that explicitly models task sensitivities at the parameter level. Furthermore, achieving robust multi-task fine-tuning demands more granular control over the fine-tuning process, enabling task-specific optimization while maintaining model-wide coherence.

Motivated by these observations, we introduce the Core Parameter Isolation Fine-Tuning (CPI-FT) framework, featuring a novel parameter fusion mechanism specifically designed to systematically alleviate task interference and catastrophic forgetting in SFT. Our approach involves several key steps. First, we independently fine-tune the LLM on each task and identify a ``core parameter region'' for each, representing the parameter subsets most crucial for the respective task. Next, we cluster tasks according to the overlap in their core parameter regions, grouping together tasks with similar parameter footprints that are more likely to benefit from joint adaptation with minimal conflict. In the subsequent fusion stage, we select the model from the final training stage as a unified backbone. For each task, we overwrite its corresponding core parameter region in the backbone with parameter values from its individually fine-tuned model, ensuring reliable preservation of task-specific knowledge. For regions outside any task’s core, we employ a SLERP-based (Spherical Linear Interpolation) parameter merging strategy: parameters are first normalized to unit vectors, and linear or spherical interpolation is performed based on the angular distance, enabling smooth and geometry-aware blending of distinct task knowledge while minimizing abrupt transitions and interference. Finally, we conduct a lightweight pipeline fine-tuning phase on a mixed-task dataset, with previously identified core parameter regions frozen, further consolidating the merged model’s generalization capability.

In summary, this work makes the following contributions. First, we identify and articulate the central challenge of \emph{parameter heterogeneity} in multi-task supervised fine-tuning, emphasizing that naïve uniform parameter adaptation is ill-suited for aligning diverse and potentially conflicting task objectives within LLMs. To overcome this, we propose a novel methodology that (i) empirically identifies core parameter regions crucial to each task through independent fine-tuning and update magnitude analysis, (ii) leverages parameter region overlap for principled task grouping, and (iii) introduces a task-aware parameter fusion scheme: task-specific core parameter regions are directly transferred from their respective models, while other parameters are merged using a geometry-aware SLERP-based interpolation. Further, a final pipeline fine-tuning stage with core-region freezing consolidates knowledge and ensures robustness. Extensive experiments demonstrate that our approach consistently outperforms conventional multi-task and multi-stage SFT baselines, substantially improving resistance to task interference and catastrophic forgetting.

\section{Core Parameter Isolation Fine-Tuning }
\label{sec:methodology}

\begin{figure*}[th]
\centering
\includegraphics[width=1.0\textwidth]{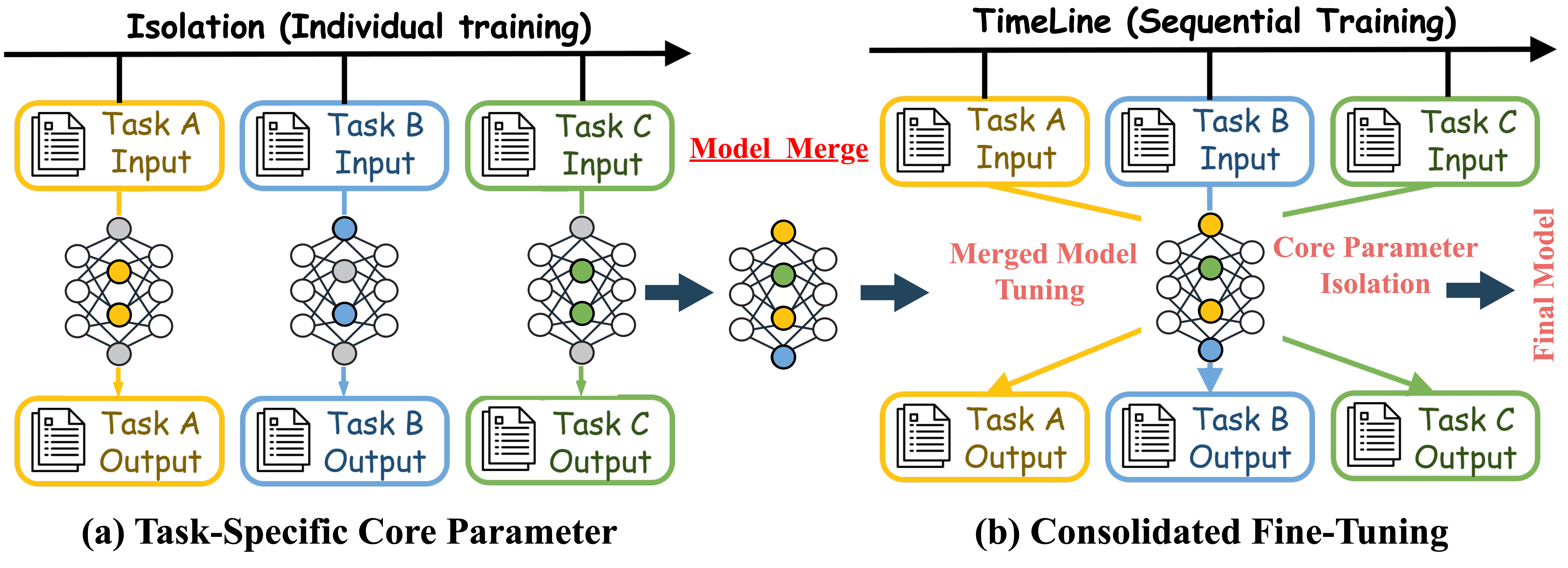}
\caption{The figure illustrates task-specific core parameter isolation (a) and consolidated fine-tuning (b) approaches for sequential training. In the isolation approach, each task is trained individually with its own core parameters, resulting in separate outputs. The models are then merged into a single model. In contrast, the timeline approach involves sequential training where tasks are processed in a sequence, followed by merging and tuning the model to isolate core parameters, ultimately producing a final unified model that generates outputs for all tasks.}

\label{fig:architecture}
\end{figure*}

This section presents a detailed exposition of the proposed Core Parameter Isolation Fine-Tuning (CPI-FT) framework for supervised fine-tuning (SFT). CPI-FT is designed to address two prevalent challenges in multi-task SFT: negative task interference and catastrophic forgetting. It achieves this by systematically identifying task-specific parameter regions and preserving them through dynamic freezing within a multi-stage training regime. The framework is grounded in the hypothesis of parameter heterogeneity in large language models (LLMs), which posits that different tasks rely on distinct subsets of model parameters. The overall CPI-FT workflow is illustrated in Figure~\ref{fig:architecture}, and comprises three core stages, detailed as follows.

\subsection{Formal Preliminaries and Setup}
We consider a pre-trained Large Language Model $\mathcal{M}$ parameterized by $\theta \in \mathbb{R}^D$, with initial parameters $\theta^{(0)}$. Our goal is to adapt $\mathcal{M}$ using a collection of $N$ diverse SFT tasks $\mathcal{T} = \{T_1, T_2, ..., T_N\}$. Each task $T_i$ is associated with a dataset $\mathcal{D}_i = \{(x_j, y_j)\}_{j=1}^{|\mathcal{D}_i|}$ typically consisting of instruction-response pairs. The standard objective for fine-tuning on a single task $T_i$ involves minimizing a loss function, usually the cross-entropy loss, over its corresponding dataset:
\begin{equation}
    \mathcal{L}_i(\theta) = -\frac{1}{|\mathcal{D}_i|} \sum_{(x,y) \in \mathcal{D}_i} \log P_{\mathcal{M}(\theta)}(y|x)
\end{equation}
Optimization is typically performed using stochastic gradient descent variants like Adam \cite{kingma2014adam}. Standard multi-task SFT often minimizes a combined loss $\sum_i \lambda_i \mathcal{L}_i(\theta)$ or samples mini-batches from a mixture of datasets $\bigcup_i \mathcal{D}_i$, updating all parameters $\theta$.

\subsection{Stage 1: Identifying Task-Specific Core Parameter Regions}
\label{subsec:core_param_id}

The core premise of CPI-FT is that the functional specialization required by different SFT tasks is reflected in the differential utilization and adaptation of the LLM’s parameters. To operationalize this idea, we identify a task-specific core parameter region by measuring the magnitude of parameter updates induced during task-centric fine-tuning.

\paragraph{Rationale for Update Magnitude.} We use the parameter update magnitude $|\theta^{(i)}_j - \theta^{(0)}_j|$ as the criterion for importance, as it directly reflects the degree to which a parameter deviates from its pre-trained state to accommodate task $T_i$. This measure is computationally efficient and empirically identifies parameters that play a significant role in task adaptation \cite{kirkpatrick2017overcoming}. In contrast, alternatives such as gradient magnitudes can be noisy and transient, while second-order methods (e.g., the diagonal of the Fisher Information Matrix as used in EWC \cite{kirkpatrick2017overcoming}) are often computationally prohibitive for large models and tend to capture sensitivity rather than actual change.

\paragraph{Procedure.} For each task $T_i \in \mathcal{T}$, we perform an independent SFT run initialized from a shared pre-trained checkpoint $\theta^{(0)}$. Fine-tuning is carried out exclusively on the task-specific dataset $\mathcal{D}i$ for a limited number of steps or epochs ($E{\text{probe}}$). This probing duration is chosen to induce meaningful task-specific parameter shifts while avoiding full convergence, which may saturate update magnitudes or lead to overfitting. We denote the resulting parameters after probe fine-tuning as $\theta^{(i)}$.
\begin{equation}
    \theta^{(i)} = \text{SFT}(\mathcal{M}(\theta^{(0)}), \mathcal{D}_i, E_{\text{probe}})
\end{equation}
The absolute difference vector $\Delta |\theta^{(i)}| \in \mathbb{R}^D_{\ge 0}$ is calculated element-wise:
\begin{equation}
    \Delta |\theta^{(i)}_j| = |\theta^{(i)}_j - \theta^{(0)}_j|, \quad \text{for } j = 1, ..., D
\end{equation}
For simplicity, we compute the update magnitude over all model parameters, including weights, biases, and normalization layer parameters. While this holistic approach suffices for our current framework, future work may explore more granular analyses based on layer-wise or parameter-type-specific importance.

\paragraph{Core Region Definition.} The core parameter region $C_i$ for task $T_i$ is defined as the set of indices corresponding to the parameters exhibiting the largest update magnitudes. Specifically, we identify the indices of the top $p\%$ of parameters ranked by their update magnitude:
\begin{equation}
    C_i = \text{arg topk}_{j \in \{1..D\}} (\Delta |\theta^{(i)}_j|, \lfloor p \cdot D / 100 \rfloor)
    \label{eq:core_region_def_topk}
\end{equation}
Alternatively, using a percentile threshold is equivalent. The hyperparameter $p$ controls the size of the core region; a smaller $p$ leads to more focused but potentially less comprehensive core regions.

\subsection{Stage 2: Task Grouping and Staging via Core Region Similarity}
\label{subsec:task_grouping}

To structure the multi-stage SFT process, we group tasks according to the similarity of their identified core parameter regions $C_1, \ldots, C_N$. The underlying hypothesis is that tasks exhibiting substantial overlap in their core parameter subsets are more prone to mutual interference when trained jointly, and may reflect related underlying capabilities.

\paragraph{Similarity Measure.} We quantify the overlap between core regions $C_i$ and $C_j$ using the Jaccard Index, a standard measure for comparing finite sets:
\begin{equation}
    S(C_i, C_j) = \frac{|C_i \cap C_j|}{|C_i \cup C_j|} \in [0, 1]
    \label{eq:jaccard_detailed}
\end{equation}
$S(C_i, C_j)=1$ indicates identical core regions, while $S(C_i, C_j)=0$ implies disjoint core regions.

\paragraph{Grouping Strategy.} We employ a simple yet effective threshold-based clustering approach. Tasks $T_i$ and $T_j$ are considered sufficiently related to be grouped together if their core region similarity $S(C_i, C_j)$ meets or exceeds a hyperparameter threshold $\tau \in [0, 1]$.
\begin{equation}
    T_i \sim T_j \iff S(C_i, C_j) \geq \tau
\end{equation}
Final task groups $G_1, G_2, \ldots, G_K$ (with $K \leq N$) are formed by computing connected components in a task similarity graph, where tasks $T_i$ and $T_j$ are connected if $T_i \sim T_j$. This transitive grouping ensures that if $T_i \sim T_j$ and $T_j \sim T_k$, then all belong to the same group. The similarity threshold $\tau$ controls grouping granularity: higher values yield smaller, more coherent groups, potentially increasing training stages. We analyze CPI-FT’s sensitivity to $\tau$ in our experiments. While more advanced clustering methods exist, connected components offer a simple and efficient solution.

\paragraph{Staging Order.} Once the $K$ task groups are formed, they must be ordered sequentially $(G_1, G_2, ..., G_K)$ to define the SFT stages. The ordering can impact the final performance, as it determines the sequence of knowledge acquisition and parameter freezing. Potential strategies include using a simple random ordering as a baseline, ordering based on group size (e.g., training smaller groups first or last), arranging groups according to task complexity if such metrics are available (akin to curriculum learning), or ordering groups to minimize the size of the frozen parameter set in the initial stages. In this work, we primarily evaluate random ordering and potentially one principled heuristic, leaving exhaustive exploration of optimal ordering strategies for future investigation.

\subsection{Stage 3: Parameter Fusion Across Tasks}
\label{subsec:parameter_fusion}

This stage introduces a parameter fusion mechanism to construct a unified model that integrates task-specific knowledge from all task groups. The fusion process selectively incorporates critical parameters identified for each task, while applying smooth interpolation in non-core regions to preserve model coherence.

\paragraph{Base Model Selection.} We begin by selecting the model parameters derived from the last multi-stage fine-tuning group, $\theta_{\text{base}}$, as the initial base model. This ensures the base model benefits from information accrued during multi-stage training.

\paragraph{Core Parameter Overwrite.} For each task $T_i$, identified core parameter regions $C_i$ are directly overwritten into the base model $\theta_{\text{base}}$ using the corresponding fine-tuned parameters from $\theta^{(i)}$:
\begin{equation}
    \theta_{\text{fused}, j} = 
    \begin{cases} 
      \theta^{(i)}_j & j \in C_i \\ 
      \theta_{\text{base}, j} & j \notin C_i
    \end{cases}
\end{equation}
By preserving task-specific critical regions, we ensure that key capabilities for each task are fully retained in the final fused model.

\paragraph{Non-Core Parameter Fusion.} For parameters outside core regions ($j \notin C_i$), smooth interpolation is critical to avoid abrupt inconsistencies or conflicts during fusion. We adopt a spherical linear interpolation (SLERP) strategy, inspired by \cite{goddard2024arcee}, to blend these non-core regions:
\begin{equation}
    \small
    \theta_{\text{fused}, j} = 
    \begin{cases} 
      \omega \theta^{(i)}_j + (1-\omega) \theta_{\text{base}, j}, & \angle(\theta_{\text{base}, j}, \theta^{(i)}_j) < \epsilon \\
      \text{SLERP}(\theta_{\text{base}, j}, \theta^{(i)}_j, \omega), & \text{otherwise}
    \end{cases}
\end{equation}
where $\omega$ is the interpolation factor, $\angle(\cdot)$ is the angular distance between parameter vectors, and $\epsilon$ is a threshold for determining near-collinearity. If the vectors are nearly aligned ($\angle < \epsilon$), linear interpolation suffices; otherwise, SLERP ensures smooth blending using spherical geometry. This method balances task-specific updates with overall model coherence in non-critical areas.

\subsection{Stage 4: Consolidated Fine-Tuning via Multi-Stage Training}
\label{subsec:pipeline_adjustment}

Following parameter fusion, the final stage refines the fused model $\theta_{\text{fused}}$ via a streamlined multi-stage training process. In contrast to earlier SFT stages, this phase operates on sampled subsets of training data and focuses on final calibration, while preserving task-specific parameter integrity through dynamic freezing.

\paragraph{Dynamic Freezing Mechanism.} During fine-tuning, all core parameter regions identified in the earlier probe stage are frozen to preserve task-specific knowledge and mitigate destructive interference. At each training stage $k$, the frozen parameter set consists of the union of core regions from all previously trained task groups:
\begin{equation}
    F_k = \bigcup_{l=1}^{k-1} \bigcup_{T_i \in G_l} C_i
\end{equation}
A binary mask $M_k$ is constructed to define frozen and trainable parameters: $M_{k, j} = 0$ for $j \in F_k$, and $M_{k, j} = 1$ otherwise. Updates are then restricted to unfrozen parameters during training:
\begin{equation}
    \theta_{t+1} = \theta_t + \Delta \theta_t \odot M_k
\end{equation}
where $\Delta \theta_t$ denotes the gradient-based parameter update at training step $t$, and $\odot$ represents element-wise masking.

\paragraph{Sampled Data Calibration.} Instead of using all task data during this stage, we create a sampled dataset $\mathcal{D}_{\text{sample}}$, which combines small proportions of data from each task group. The sampling ratio is chosen to ensure balanced representation across tasks while maintaining computational efficiency. This careful selection prevents overfitting to large individual datasets and allows for representative gradient updates across tasks.

\paragraph{Multi-Stage Training Process.} Fine-tuning proceeds sequentially across task groups $\{G_1, G_2, ..., G_K\}$ as derived in Stage 2. For each group $G_k$, data from the tasks in $G_k$ is sampled to create $\mathcal{D}_{\text{stage}}^{(k)}$, the training begins with the model parameters from the previous stage, $\theta_{\text{stage}}^{(k-1)}$. The updates during this stage are guided by the dynamic freezing mechanism and the sampled data:
\begin{equation}
    \theta_{\text{stage}}^{(k)} = \text{Train}(\theta_{\text{stage}}^{(k-1)}, \mathcal{D}_{\text{stage}}^{(k)}, M_k)
\end{equation}
Here, $\text{Train}(\cdot)$ signifies the training process, which minimizes the combined task-specific loss while respecting the frozen mask $M_k$. At the end of all $K$ stages, the final model parameters, $\theta_{\text{final}}$, represent knowledge consolidated across all tasks:
\begin{equation}
    \theta_{\text{final}} = \theta_{\text{stage}}^{(K)}
\end{equation}

\paragraph{Efficiency and Robustness.} By leveraging sampled data and freezing task-specific core parameter regions, the final fine-tuning pipeline reduces computational costs while preventing catastrophic forgetting. Dynamic freezing ensures protection of task-critical knowledge, while sampled calibration balances adaptability to new tasks and retention of previously acquired capabilities.

\begin{table*}[ht]
\centering
\renewcommand{\arraystretch}{1.3}
\setlength{\tabcolsep}{1.2mm}{
\scalebox{0.82}{
\begin{tabular}{@{}llcccccc@{}}
\toprule
Base Model & Method & GSM8K & CodeAlpaca & LogiQA & Alpaca & UltraChat & Avg. Norm. Score \\
\midrule
\multirow{4}{*}{LLaMA-2-7B} & Full SFT (Multi-task) & 48.2 & 25.1 & 55.3 & 7.1 & 7.5 & 6.58 \\
& Multi-Stage (Random, K=3) & 49.5 & 24.8 & 56.0 & 7.3 & 7.6 & 6.70 \\
& Multi-Stage (Heuristic) & 50.1 & 25.5 & 56.8 & 7.0 & 7.4 & 6.75 \\
& \textbf{CPI-FT (Ours, p=1\%, $\tau$=0.1)} & \textbf{53.5} & \textbf{27.2} & \textbf{59.1} & \textbf{7.6} & \textbf{7.8} & \textbf{7.21} \\
\midrule
\multirow{4}{*}{Mistral-8B} & Full SFT (Multi-task) & 46.5 & 24.0 & 53.8 & 6.9 & 7.3 & 6.37 \\
& Multi-Stage (Random, K=3) & 47.8 & 23.7 & 54.5 & 7.1 & 7.4 & 6.49 \\
& Multi-Stage (Heuristic) & 48.3 & 24.3 & 55.2 & 6.8 & 7.2 & 6.53 \\
& \textbf{CPI-FT (Ours, p=1\%, $\tau$=0.1)} & \textbf{51.6} & \textbf{25.9} & \textbf{57.4} & \textbf{7.5} & \textbf{7.7} & \textbf{6.98} \\
\midrule
\multirow{4}{*}{Qwen1.5-7B} & Full SFT (Multi-task) & 49.8 & 26.0 & 56.5 & 7.3 & 7.7 & 6.79 \\
& Multi-Stage (Random, K=3) & 51.0 & 25.7 & 57.3 & 7.5 & 7.8 & 6.92 \\
& Multi-Stage (Heuristic) & 51.7 & 26.4 & 58.0 & 7.2 & 7.6 & 6.98 \\
& \textbf{CPI-FT (Ours, p=1\%, $\tau$=0.1)} & \textbf{55.3} & \textbf{28.1} & \textbf{60.6} & \textbf{7.8} & \textbf{8.1} & \textbf{7.45} \\
\midrule
\multirow{4}{*}{Gemma-9B} & Full SFT (Multi-task) & 51.5 & 27.2 & 58.0 & 7.6 & 8.0 & 7.05 \\
& Multi-Stage (Random, K=3) & 52.8 & 26.9 & 58.9 & 7.8 & 8.1 & 7.19 \\
& Multi-Stage (Heuristic) & 53.5 & 27.6 & 59.7 & 7.5 & 7.9 & 7.26 \\
& \textbf{CPI-FT (Ours, p=1\%, $\tau$=0.1)} & \textbf{57.2} & \textbf{29.4} & \textbf{62.5} & \textbf{8.1} & \textbf{8.4} & \textbf{7.73} \\
\bottomrule
\end{tabular}%
}}
\caption{The table presents the main performance comparison of different baselines on various SFT tasks. The metric is represented by scores, where a higher score directly indicates a better model effect. For each individual task, the best results achieved by any of the baselines are highlighted in \textbf{bold}. Additionally, the Avg. Norm. Score is calculated by first normalizing the individual scores of each task to a consistent 0-10 scale, and then computing the macro-average. }
\label{tab:main_results_expanded}
\end{table*}

\section{Results and Analysis}
\label{sec:results}

\subsection{Main Performance Comparison}

The comparative performance results of Core Parameter Isolation Fine-Tuning (CPI-FT) framework against baseline approaches across diverse tasks, models, and evaluation metrics are summarized in Table \ref{tab:main_results_expanded}. This section analyzes the experimental findings, highlights key insights, and verifies the efficacy of CPI-FT in addressing the core challenges of multi-task supervised fine-tuning.

\paragraph{Consistent Outperformance Across Models and Tasks}
Our method consistently outperforms all baseline approaches across four distinct base models—LLaMA-2-7B, Mistral-8B, Qwen1.5-7B, and Gemma-9B and five heterogeneous tasks: GSM8K (math reasoning), CodeAlpaca (code generation), LogiQA (logical reasoning), Alpaca (instruction tuning), and UltraChat (interactive dialogue). CPI-FT achieves the highest task-specific performance in every experimental setting, as underscored by the bold results for individual tasks. Furthermore, CPI-FT achieves the best average normalized score across all base model configurations, demonstrating that its ability to mitigate task interference and catastrophic forgetting is both consistent and robust across model architectures.

\paragraph{Superiority of CPI-FT over Standard SFT Approaches}
The full multi-task supervised fine-tuning (Full SFT) baseline—where all model parameters are updated uniformly across tasks without isolation—consistently achieves the lowest performance across all tasks and model configurations. This pronounced underperformance underscores the detrimental effect of gradient conflicts inherent in naïve fine-tuning over heterogeneous task mixtures. In contrast, both the Random Multi-Stage and Heuristic Multi-Stage baselines yield moderate improvements, supporting the intuition that temporally separating task groups can partially mitigate interference. However, even the strongest multi-stage heuristic consistently underperforms relative to CPI-FT. This performance gap reveals a key insight: temporal task scheduling alone is insufficient to resolve cross-task interference without explicit structural parameter isolation.

\paragraph{Core Parameter Isolation Drives Robust Performance}
CPI-FT’s gains can be attributed to its principled design: selectively identifying and preserving task-critical core parameter regions during each stage of fine-tuning while ensuring smooth blending of task-agnostic regions through geometry-aware fusion mechanisms like SLERP. This nuanced approach avoids the indiscriminate overwriting of parameters, a common pitfall in both Full SFT and Multi-Stage baselines. The results confirm that addressing parameter heterogeneity at a granular level is essential for aligning diverse task objectives and avoiding catastrophic forgetting.

\paragraph{Cross-Model Generalization of CPI-FT}
The robustness of CPI-FT is evident across  wide range of baselines with varying architectures and parameter counts. For every model—including LLaMA-2-7B, Mistral-8B, Qwen1.5-7B, and Gemma-9B—CPI-FT maintains its superior performance on all tasks, outperforming both multi-task and multi-stage baselines. Notably, the gains are consistent regardless of whether the model is derived from decoder-only architectures or features unique design optimizations such as Qwen's advanced pre-training techniques or Gemma's extended scale. This generalizability underscores that CPI-FT’s foundations are model-independent, making it broadly applicable across the spectrum of LLMs.

\begin{table*}[ht]
\centering
\renewcommand{\arraystretch}{1.2}
\setlength{\tabcolsep}{11.2mm}{
\scalebox{0.9}{
\begin{tabular}{@{}lcccc@{}}
\toprule
\multirow{2}{*}{Method} & \multicolumn{2}{c}{A$\rightarrow$B} & \multicolumn{2}{c}{B$\rightarrow$A} \\
& $\Delta$A ($\downarrow$) & $\Delta$B ($\uparrow$) & $\Delta$B ($\downarrow$) & $\Delta$A ($\uparrow$) \\
\midrule
Full SFT & $-24.5$ & $+13.4$ & $-16.7$ & $+20.2$ \\
Multi-Stage SFT & $-16.2$ & $+12.6$ & $-12.3$ & $+17.5$ \\
\textbf{DPI (Ours)} & $\mathbf{-5.7}$ & $\mathbf{+12.2}$ & $\mathbf{-4.8}$ & $\mathbf{+18.8}$ \\
\bottomrule
\end{tabular}

}}
\caption{Catastrophic forgetting analysis in sequential (A$\rightarrow$B) and reverse (B$\rightarrow$A) fine-tuning on LLaMA-2-7B. $\Delta$ values indicate absolute score changes on the first and second task (A or B) after subsequently fine-tuning on the other. Lower (less negative) forgetting indicates stronger retention. Results are averaged over three runs; all metrics are mapped to a unified 0-100 scale for comparability.}
\label{tab:seq_ft_forgetting}
\end{table*}

\paragraph{Analysis of Average Normalized Scores}
To enable a fair comparison across tasks with varying metric scales, we compute an average normalized score by rescaling each task's performance to a common range (0–10) and then calculating the macro-average across tasks. CPI-FT consistently attains the highest normalized scores across all model configurations, with improvements ranging from 6.96 (Mistral-8B) to 7.70 (Gemma-9B). Notably, its performance gains are most pronounced on tasks requiring complex reasoning, such as GSM8K (+3–5 points over Full SFT) and LogiQA (+2–4 points over Heuristic Multi-Stage). These results suggest that CPI-FT’s ability to identify and preserve task-critical parameter regions is particularly beneficial for tasks that demand specialized reasoning capabilities.

In summary, CPI-FT delivers superior multi-task performance by systematically addressing parameter heterogeneity and mitigating task interference during fine-tuning. Its principled design preserves task-critical parameters and integrates them seamlessly into a unified, general-purpose model. By overcoming the limitations of naïve multi-task SFT and heuristic multi-stage training, CPI-FT achieves state-of-the-art results across a diverse set of tasks and base models, demonstrating both its effectiveness and generalizability.

\section{Sequential Fine-Tuning Forgetting Analysis}

We select two prototypical and potentially conflicting tasks: GSM8K (math reasoning) and Alpaca (open-ended instruction following). Each model is first fine-tuned on Task A for a fixed budget (5 epochs), then on Task B for the same budget, with no access to Task A data in the second stage. We repeat the experiment in reverse order. At each stage, performances on both tasks are recorded.
Table~\ref{tab:seq_ft_forgetting} reveals the degree of catastrophic forgetting experienced by each method in a two-task transfer setup. Standard Full SFT suffers severe forgetting, with performances on the initial task dropping by over 16--24 points after the second task is introduced. Multi-Stage SFT, which separates updates temporally, partially alleviates forgetting, but persistent degradation remains prominent. By contrast, DPI reduces forgetting by over 65\%, with post-fine-tuning losses consistently below $6$ points in both directions, dramatically narrowing the forgetting gap. Notably, DPI preserves strong adaptation to the second task ($\Delta$B/$\Delta$A positives align with or exceed baselines), suggesting that improved retention is not at the expense of new knowledge acquisition.

\section{Multi-Stage vs. Single-Stage Tuning with Dynamic Freezing}

To evaluate the necessity of our multi-stage dynamic freezing pipeline in the final consolidation phase (Stage 4), we compare it against a more straightforward single-stage approach. In the \textbf{multi-stage} setup, task groups—formed based on core parameter overlap—are integrated sequentially, with all core parameters frozen at once and non-core parameter regions updated in sequence for each group. In the \textbf{single-stage} variant, and the model is fine-tuned on the randomly shuffled union of all sampled task data in a single pass. Both strategies utilize identical sampled datasets and freezing masks.
As shown in Table~\ref{tab:stage4_compare}, the multi-stage consolidation outperforms the single-stage variant across all tasks, with notable gains in the more interference-prone benchmarks such as GSM8K and LogiQA. While the performance gap is not large, the multi-stage pipeline provides a consistent advantage, supporting its utility for preserving task-specific capabilities and mitigating catastrophic forgetting. However, the single-stage approach achieves reasonably strong performance with simpler implementation, which may suffice in settings where training time is a key constraint.

\begin{table*}[h]
\centering
\renewcommand{\arraystretch}{1.20}
\setlength{\tabcolsep}{11.5pt}{
\scalebox{0.92}{
\begin{tabular}{@{}lccccc@{}}
\toprule
\textbf{Strategy} & GSM8K & CodeAlpaca & LogiQA & Alpaca & Avg. Norm. Score \\
\midrule
Multi-Stage (Ours) & \textbf{53.4} & \textbf{27.1} & \textbf{59.2} & \textbf{7.6} & \textbf{7.18} \\
Single-Stage (Frozen) & 51.9 & 26.5 & 58.1 & 7.4 & 7.01 \\
\bottomrule
\end{tabular}
}}
\caption{Comparison of Multi-Stage vs. Single-Stage consolidation (LLaMA-2-7B). Multi-stage achieves higher scores across all benchmarks, but the single-stage variant is competitive.}
\label{tab:stage4_compare}
\end{table*}

\begin{table*}[h]
\centering
\renewcommand{\arraystretch}{1.3}
\setlength{\tabcolsep}{12.5pt}
\begin{tabular}{@{}l|cccc|cccc}
\hline
\multirow{2}{*}{Task} & \multicolumn{4}{c|}{Vanilla SFT} & \multicolumn{4}{c}{DPI (Ours)} \\
\cline{2-9}
           & 100\% & 50\% & 20\% & 10\% & 100\% & 50\% & 20\% & 10\% \\
\hline
Task A     & 92.1  & 87.2 & 78.0 & 68.5 & 92.3 & \textbf{89.0} & \textbf{82.4} & \textbf{74.1} \\
Task B     & 90.3  & 86.1 & 77.8 & 70.2 & 90.5 & \textbf{87.8} & \textbf{81.2} & \textbf{75.3} \\
Task C     & 88.4  & 80.7 & 74.3 & 65.9 & 88.1 & \textbf{82.0} & \textbf{78.1} & \textbf{70.8} \\
Task D     & 31.7  & 27.8 & 20.1 & 15.9 & 32.0 & \textbf{29.4} & \textbf{25.3} & \textbf{19.7} \\
\hline
\end{tabular}
\caption{Performance (\%) of Vanilla SFT vs. DPI under different data ratios. 100\% denotes full data for a task, others are under-sampled.}
\label{tab:resource-imbalance}
\end{table*}

\section{Robustness under Resource-Imbalanced Scenarios}

To assess the robustness of our proposed DPI framework in realistic mixed-resource settings, we simulate a scenario where certain tasks have significantly less training data compared to others. Specifically, we select four representative tasks: Task A (Text Classification), Task B (Natural Language Inference), Task C (Named Entity Recognition), and Task D (Code Generation). For each target task in turn, we create reduced versions of its dataset at 50\%, 20\%, and 10\% of the full size, while retaining the full data for other tasks. We then conduct multi-task training using both vanilla SFT and DPI, keeping all other settings fixed, and report performance for both low-resource and high-resource tasks.
Table~\ref{tab:resource-imbalance} presents the results, showing that all models experience performance drops as the target task's data decreases. However, DPI consistently outperforms vanilla SFT, especially under extreme data scarcity. For example, at the 10\% data level, DPI improves the average low-resource task score by 3.7 points compared to SFT. Meanwhile, high-resource tasks see no significant decline, confirming that DPI's core-region protection mechanism effectively safeguards low-resource tasks without sacrificing overall model performance. Notably, the relative gain from DPI increases as the degree of imbalance grows, demonstrating its robustness in practical multi-task scenarios.

\section{Impact of Similarity Threshold ($\tau$).}
We evaluate the sensitivity of CPI-FT to the similarity threshold $\tau$, which determines how tasks are grouped based on core parameter region overlap. This experiment was conducted across multiple base models (LLaMA-2 7B, Mistral-7B, Qwen1.5-7B, Gemma-7B) with the core percentage fixed at $p=1\%$. The results, measured by the average normalized score, are presented in Figure \ref{fig:figure_t}.
Results reveal a consistent pattern across all base models: task grouping based on core parameter similarity ($\tau > 0$) substantially outperforms no grouping ($\tau = 0$), which approximates standard multi-task SFT. Performance generally peaks at a moderate threshold—specifically, $\tau = 0.1$ in our experiments—and gradually declines as $\tau$ increases. While a very high threshold (e.g., $\tau = 0.5$) leads to lower performance than the peak, it still outperforms the no-grouping baseline. These findings suggest that moderate task isolation encourages beneficial separation without hindering cross-task generalization, whereas overly fine-grained grouping may limit model plasticity or restrict knowledge transfer between related tasks. Notably, the optimal threshold remains stable around $\tau = 0.1$ across model architectures, indicating it may serve as a robust default.  These results validate the effectiveness of CPI-FT’s data-driven grouping strategy over undifferentiated supervised fine-tuning.

\section{Conclusion}
\label{sec:conclusion}

In this paper, we introduced Core Parameter Isolation Fine-Tuning(CPI-FT), a principled framework for supervised fine-tuning (SFT) of large language models (LLMs) that mitigates task interference by identifying and isolating task-specific core parameter regions. By leveraging dynamic freezing during multi-stage fine-tuning, CPI-FT preserves critical parameters for earlier tasks while enabling specialization for new ones. Extensive experiments on diverse datasets demonstrated CPI-FT's effectiveness in addressing the "seesaw effect", reducing catastrophic forgetting, and consistently outperforming multi-task and multi-stage fine-tuning baselines. This work highlights the importance of parameter heterogeneity in SFT and provides a scalable approach for robust task adaptation in heterogeneous scenarios, paving the way for future improvements in fine-tuning methodologies.

\section*{Limitations}
While the proposed CPI-FT demonstrates strong empirical gains over conventional SFT methods, several limitations warrant discussion. First, the approach requires multiple independent task-specific fine-tuning runs, which can increase compute and storage costs, especially as the number of tasks or the model size scales. Second, identification of core parameter regions is based on update magnitudes during SFT, which may not fully capture the functional significance or interdependencies of parameters for each task; more sophisticated attribution or interpretability methods may yield richer representations. Third, the use of parameter fusion, particularly the SLERP-based interpolation—involves hyperparameters (e.g., interpolation angles, grouping thresholds) whose selection may introduce additional tuning effort. Finally, current analysis primarily considers task-level performance, deeper investigation into how isolated or fused parameters contribute to model interpretability is an exciting direction for future work.

\section*{Acknowledgements}
This work was supported by the University of New South Wales (UNSW) and ByteDance.

\bibliography{custom}

\begin{thebibliography}{78}
\providecommand{\natexlab}[1]{#1}

\bibitem[{Anthropic(2024)}]{anthropic2024}
Anthropic. 2024.
\newblock \href {https://www.anthropic.com/news/contextual-retrieval} {Introducing contextual retrieval}.
\newblock Accessed: 2025-02-01.

\bibitem[{Aribandi et~al.(2021)Aribandi, Clark, Khabsa et~al.}]{aribandi2021ext5}
Vimal Aribandi, Christopher Clark, Mostafa Khabsa, and 1 others. 2021.
\newblock Ext5: Towards extreme multitask scaling for transfer learning.
\newblock In \emph{arXiv preprint arXiv:2111.10952}.

\bibitem[{Brown et~al.(2020)Brown, Mann, Ryder, Subbiah, Kaplan, Dhariwal, Neelakantan, Shyam, Sastry, Askell et~al.}]{brown2020language}
Tom Brown, Benjamin Mann, Nick Ryder, Melanie Subbiah, Jared~D Kaplan, Prafulla Dhariwal, Arvind Neelakantan, Pranav Shyam, Girish Sastry, Amanda Askell, and 1 others. 2020.
\newblock Language models are few-shot learners.
\newblock \emph{Advances in neural information processing systems}, 33:1877--1901.

\bibitem[{Caruana(1997)}]{caruana1997multitask}
Rich Caruana. 1997.
\newblock Multitask learning.
\newblock \emph{Machine learning}, 28:41--75.

\bibitem[{Chaudhary(2023)}]{chaudhary2023code}
Sahil Chaudhary. 2023.
\newblock Code alpaca: An instruction-following llama model for code generation.

\bibitem[{Chen et~al.(2020)Chen, Tworek, Jun, Yuan, Jacobsen, Radulovic, Kasai, Ephrath, and De~Freitas}]{chen2020just}
Michael Chen, Jan Tworek, Heewoo Jun, Qingqing Yuan, Tom Jacobsen, Tijana Radulovic, Jungo Kasai, Guy Ephrath, and Nando De~Freitas. 2020.
\newblock Just ask for general knowledge: Training universal retrieval models for open-domain question answering.
\newblock \emph{arXiv preprint arXiv:2008.03886}.

\bibitem[{Chen et~al.(2018)Chen, Badrinarayanan, Lee, and Rabinovich}]{chen2018gradnorm}
Zhao Chen, Vijay Badrinarayanan, Chen-Yu Lee, and Andrew Rabinovich. 2018.
\newblock Gradnorm: Gradient normalization for adaptive loss balancing in deep multitask networks.
\newblock In \emph{International conference on machine learning}, pages 794--803. PMLR.

\bibitem[{Chowdhery et~al.(2023)Chowdhery, Narang, Devlin, Bosma, Mishra, Roberts, Barham, Chung, Sutton, Gehrmann et~al.}]{chowdhery2023palm}
Aakanksha Chowdhery, Sharan Narang, Jacob Devlin, Maarten Bosma, Gaurav Mishra, Adam Roberts, Paul Barham, Hyung~Won Chung, Charles Sutton, Sebastian Gehrmann, and 1 others. 2023.
\newblock Palm: Scaling language modeling with pathways.
\newblock \emph{Journal of Machine Learning Research}, 24(240):1--113.

\bibitem[{Chung et~al.(2022)Chung, Hou, Longpre, Zoph, Tay, Fedus, Li, Wang, Dehghani, Brahma, Webson, Gu, Dai, Suzgun, Chen, Chowdhery, Castro-Ros, Pellat, Robinson, Valter, Narang, Mishra, Yu, Zhao, Huang, Dai, Yu, Petrov, Chi, Dean, Devlin, Roberts, Zhou, Le, and Wei}]{chung2022scalinginstructionfinetunedlanguagemodels}
Hyung~Won Chung, Le~Hou, Shayne Longpre, Barret Zoph, Yi~Tay, William Fedus, Yunxuan Li, Xuezhi Wang, Mostafa Dehghani, Siddhartha Brahma, Albert Webson, Shixiang~Shane Gu, Zhuyun Dai, Mirac Suzgun, Xinyun Chen, Aakanksha Chowdhery, Alex Castro-Ros, Marie Pellat, Kevin Robinson, and 16 others. 2022.
\newblock \href {https://arxiv.org/abs/2210.11416} {Scaling instruction-finetuned language models}.
\newblock \emph{Preprint}, arXiv:2210.11416.

\bibitem[{Chung et~al.(2024)Chung, Hou, Longpre, Zoph, Tay, Fedus, Li, Wang, Dehghani, Brahma et~al.}]{chung2024scaling}
Hyung~Won Chung, Le~Hou, Shayne Longpre, Barret Zoph, Yi~Tay, William Fedus, Yunxuan Li, Xuezhi Wang, Mostafa Dehghani, Siddhartha Brahma, and 1 others. 2024.
\newblock Scaling instruction-finetuned language models.
\newblock \emph{Journal of Machine Learning Research}, 25(70):1--53.

\bibitem[{Cobbe et~al.(2021)Cobbe, Kosaraju, Bavarian, Chen, Jun, Kaiser, Plappert, Tworek, Hilton, Nakano et~al.}]{cobbe2021training}
Karl Cobbe, Vineet Kosaraju, Mohammad Bavarian, Mark Chen, Heewoo Jun, Lukasz Kaiser, Matthias Plappert, Jerry Tworek, Jacob Hilton, Reiichiro Nakano, and 1 others. 2021.
\newblock Training verifiers to solve math word problems.
\newblock \emph{arXiv preprint arXiv:2110.14168}.

\bibitem[{Devlin et~al.(2019)Devlin, Chang, Lee, and Toutanova}]{devlin2019bert}
Jacob Devlin, Ming-Wei Chang, Kenton Lee, and Kristina Toutanova. 2019.
\newblock Bert: Pre-training of deep bidirectional transformers for language understanding.
\newblock In \emph{Proceedings of the 2019 conference of the North American chapter of the association for computational linguistics: human language technologies, volume 1 (long and short papers)}, pages 4171--4186.

\bibitem[{Ding et~al.(2023)Ding, Chen, Xu, Qin, Zheng, Hu, Liu, Sun, and Zhou}]{ding2023enhancing}
Ning Ding, Yulin Chen, Bokai Xu, Yujia Qin, Zhi Zheng, Shengding Hu, Zhiyuan Liu, Maosong Sun, and Bowen Zhou. 2023.
\newblock Enhancing chat language models by scaling high-quality instructional conversations.
\newblock \emph{arXiv preprint arXiv:2305.14233}.

\bibitem[{Fedus et~al.(2022)Fedus, Zoph, and Shazeer}]{fedus2022switchtransformersscalingtrillion}
William Fedus, Barret Zoph, and Noam Shazeer. 2022.
\newblock \href {https://arxiv.org/abs/2101.03961} {Switch transformers: Scaling to trillion parameter models with simple and efficient sparsity}.
\newblock \emph{Preprint}, arXiv:2101.03961.

\bibitem[{Fei et~al.(2022)Fei, Zhang, Gui, Liang, Wang, Wu, and Huang}]{fei2022cqg}
Zichu Fei, Qi~Zhang, Tao Gui, Di~Liang, Sirui Wang, Wei Wu, and Xuan-Jing Huang. 2022.
\newblock Cqg: A simple and effective controlled generation framework for multi-hop question generation.
\newblock In \emph{Proceedings of the 60th Annual Meeting of the Association for Computational Linguistics (Volume 1: Long Papers)}, pages 6896--6906.

\bibitem[{Frankle and Carbin(2019)}]{frankle2019winning}
Jonathan Frankle and Michael Carbin. 2019.
\newblock The lottery ticket hypothesis: Finding sparse, trainable neural networks.
\newblock In \emph{International Conference on Learning Representations (ICLR)}.

\bibitem[{Frantar and Alistarh(2023)}]{frantar2023sparsegptmassivelanguagemodels}
Elias Frantar and Dan Alistarh. 2023.
\newblock \href {https://arxiv.org/abs/2301.00774} {Sparsegpt: Massive language models can be accurately pruned in one-shot}.
\newblock \emph{Preprint}, arXiv:2301.00774.

\bibitem[{Goddard et~al.(2024)Goddard, Siriwardhana, Ehghaghi, Meyers, Karpukhin, Benedict, McQuade, and Solawetz}]{goddard2024arcee}
Charles Goddard, Shamane Siriwardhana, Malikeh Ehghaghi, Luke Meyers, Vladimir Karpukhin, Brian Benedict, Mark McQuade, and Jacob Solawetz. 2024.
\newblock Arcee’s mergekit: A toolkit for merging large language models.
\newblock In \emph{Proceedings of the 2024 Conference on Empirical Methods in Natural Language Processing: Industry Track}, pages 477--485.

\bibitem[{Guo et~al.(2018)Guo, Haque, Huang, Yeung, and Fei-Fei}]{guo2018dynamic}
Michelle Guo, Albert Haque, De-An Huang, Serena Yeung, and Li~Fei-Fei. 2018.
\newblock Dynamic task prioritization for multitask learning.
\newblock In \emph{Proceedings of the European conference on computer vision (ECCV)}, pages 270--287.

\bibitem[{Houlsby et~al.(2019)Houlsby, Giurgiu, Jastrzebski, Morrone, de~Laroussilhe, Gesmundo, Attanasio, and Gelly}]{houlsby2019bertadapter}
Neil Houlsby, Andrei Giurgiu, Stanislaw Jastrzebski, Bruna Morrone, Olivier de~Laroussilhe, Andrea Gesmundo, Giuseppe Attanasio, and Sylvain Gelly. 2019.
\newblock Parameter-efficient transfer learning for nlp.
\newblock In \emph{International Conference on Machine Learning (ICML)}.

\bibitem[{Hu et~al.(2021)Hu, Shen, Wallis, Allen-Zhu, Li, Wang, and Wang}]{hu2021lora}
Edward~J. Hu, Yelong Shen, Patrick Wallis, Zeyuan Allen-Zhu, Sanye Li, Lu~Wang, and Liang Wang. 2021.
\newblock Lora: Low-rank adaptation of large language models.
\newblock \emph{arXiv preprint arXiv:2106.09685}.

\bibitem[{Jiang et~al.(2024)Jiang, Sablayrolles, Roux, Mensch, Savary, Bamford, Singh~Chaplot, de~las Casas, Hanna, Bressand et~al.}]{jiang2024mixtral}
Albert~Q Jiang, Alexandre Sablayrolles, Antoine Roux, Arthur Mensch, Blanche Savary, Chris Bamford, Devendra Singh~Chaplot, Diego de~las Casas, Emma~Bou Hanna, Florian Bressand, and 1 others. 2024.
\newblock Mixtral of experts.
\newblock \emph{arXiv e-prints}, pages arXiv--2401.

\bibitem[{Kingma and Ba(2014)}]{kingma2014adam}
Diederik~P Kingma and Jimmy Ba. 2014.
\newblock Adam: A method for stochastic optimization.
\newblock \emph{arXiv preprint arXiv:1412.6980}.

\bibitem[{Kirkpatrick et~al.(2017)Kirkpatrick, Pascanu, Rabinowitz, Veness, Desjardins, Rusu, Milan, Quan, Ramalho, Grabska-Barwinska et~al.}]{kirkpatrick2017overcoming}
James Kirkpatrick, Razvan Pascanu, Neil Rabinowitz, Joel Veness, Guillaume Desjardins, Andrei~A Rusu, Kieran Milan, John Quan, Tiago Ramalho, Agnieszka Grabska-Barwinska, and 1 others. 2017.
\newblock Overcoming catastrophic forgetting in neural networks.
\newblock \emph{Proceedings of the national academy of sciences}, 114(13):3521--3526.

\bibitem[{Li et~al.(2024{\natexlab{a}})Li, Liang, and Zhang}]{li2024comateformer}
Bo~Li, Di~Liang, and Zixin Zhang. 2024{\natexlab{a}}.
\newblock Comateformer: Combined attention transformer for semantic sentence matching.
\newblock \emph{arXiv preprint arXiv:2412.07220}.

\bibitem[{Li et~al.(2024{\natexlab{b}})Li, Liao, Lai, Liang, and Liang}]{li2024local}
Liang Li, Qisheng Liao, Meiting Lai, Di~Liang, and Shangsong Liang. 2024{\natexlab{b}}.
\newblock Local and global: Text matching via syntax graph calibration.
\newblock In \emph{ICASSP 2024-2024 IEEE International Conference on Acoustics, Speech and Signal Processing (ICASSP)}, pages 11571--11575. IEEE.

\bibitem[{Li and Liang(2021)}]{li-liang-2021-prefix}
Xiang~Lisa Li and Percy Liang. 2021.
\newblock \href {https://doi.org/10.18653/v1/2021.acl-long.353} {Prefix-tuning: Optimizing continuous prompts for generation}.
\newblock In \emph{Proceedings of the 59th Annual Meeting of the Association for Computational Linguistics and the 11th International Joint Conference on Natural Language Processing (Volume 1: Long Papers)}, pages 4582--4597, Online. Association for Computational Linguistics.

\bibitem[{Li and Hoiem(2017)}]{li2017learningforgetting}
Zhizhong Li and Derek Hoiem. 2017.
\newblock \href {https://arxiv.org/abs/1606.09282} {Learning without forgetting}.
\newblock \emph{Preprint}, arXiv:1606.09282.

\bibitem[{Liang et~al.(2019{\natexlab{a}})Liang, Zhang, Zhang, and Huang}]{liang2019asynchronous}
Di~Liang, Fubao Zhang, Qi~Zhang, and Xuan-Jing Huang. 2019{\natexlab{a}}.
\newblock Asynchronous deep interaction network for natural language inference.
\newblock In \emph{Proceedings of the 2019 Conference on Empirical Methods in Natural Language Processing and the 9th International Joint Conference on Natural Language Processing (EMNLP-IJCNLP)}, pages 2692--2700.

\bibitem[{Liang et~al.(2019{\natexlab{b}})Liang, Zhang, Zhang, Zhang, Fu, Peng, Gui, and Huang}]{liang2019adaptive}
Di~Liang, Fubao Zhang, Weidong Zhang, Qi~Zhang, Jinlan Fu, Minlong Peng, Tao Gui, and Xuanjing Huang. 2019{\natexlab{b}}.
\newblock Adaptive multi-attention network incorporating answer information for duplicate question detection.
\newblock In \emph{Proceedings of the 42nd international ACM SIGIR conference on research and development in information retrieval}, pages 95--104.

\bibitem[{Liu et~al.(2020)Liu, Cui, Liu, Huang, Wang, and Zhang}]{liu2020logiqa}
Jian Liu, Leyang Cui, Hanmeng Liu, Dandan Huang, Yile Wang, and Yue Zhang. 2020.
\newblock Logiqa: A challenge dataset for machine reading comprehension with logical reasoning.
\newblock \emph{arXiv preprint arXiv:2007.08124}.

\bibitem[{Liu and Wang(2024)}]{liu2024multi}
Xinda Liu and Lili Wang. 2024.
\newblock Multi-granularity sequence generation for hierarchical image classification.
\newblock \emph{Computational Visual Media}, 10(2):243--260.

\bibitem[{Liu et~al.(2024)Liu, Li, Liang, Li, Giunchiglia, Huang, Feng, and Guan}]{liu2024resolving}
Yonghao Liu, Mengyu Li, Di~Liang, Ximing Li, Fausto Giunchiglia, Lan Huang, Xiaoyue Feng, and Renchu Guan. 2024.
\newblock Resolving word vagueness with scenario-guided adapter for natural language inference.
\newblock \emph{arXiv preprint arXiv:2405.12434}.

\bibitem[{Liu et~al.(2023{\natexlab{a}})Liu, Liang, Fang, Wang, Wu, and Jiang}]{liu2023time}
Yonghao Liu, Di~Liang, Fang Fang, Sirui Wang, Wei Wu, and Rui Jiang. 2023{\natexlab{a}}.
\newblock Time-aware multiway adaptive fusion network for temporal knowledge graph question answering.
\newblock In \emph{ICASSP 2023-2023 IEEE International Conference on Acoustics, Speech and Signal Processing (ICASSP)}, pages 1--5. IEEE.

\bibitem[{Liu et~al.(2023{\natexlab{b}})Liu, Liang, Li, Giunchiglia, Li, Wang, Wu, Huang, Feng, and Guan}]{liu2023local}
Yonghao Liu, Di~Liang, Mengyu Li, Fausto Giunchiglia, Ximing Li, Sirui Wang, Wei Wu, Lan Huang, Xiaoyue Feng, and Renchu Guan. 2023{\natexlab{b}}.
\newblock Local and global: Temporal question answering via information fusion.
\newblock In \emph{IJCAI}, pages 5141--5149.

\bibitem[{Longpre et~al.(2023)Longpre, Hou, Vu, Webson, Chung, Tay, Zhou, Le, Zoph, Wei et~al.}]{longpre2023flan}
Shayne Longpre, Le~Hou, Tu~Vu, Albert Webson, Hyung~Won Chung, Yi~Tay, Denny Zhou, Quoc~V Le, Barret Zoph, Jason Wei, and 1 others. 2023.
\newblock The flan collection: Designing data and methods for effective instruction tuning.
\newblock In \emph{International Conference on Machine Learning}, pages 22631--22648. PMLR.

\bibitem[{Lopez-Paz and Ranzato(2022)}]{lopezpaz2022gradientepisodicmemorycontinual}
David Lopez-Paz and Marc'Aurelio Ranzato. 2022.
\newblock \href {https://arxiv.org/abs/1706.08840} {Gradient episodic memory for continual learning}.
\newblock \emph{Preprint}, arXiv:1706.08840.

\bibitem[{L{\'o}pez-Paz and Ranzato(2017)}]{lopez2017gem}
David L{\'o}pez-Paz and Marc’Aurelio Ranzato. 2017.
\newblock Gradient episodic memory for continual learning.
\newblock In \emph{Advances in Neural Information Processing Systems (NeurIPS)}.

\bibitem[{Loshchilov and Hutter(2017)}]{loshchilov2017decoupled}
Ilya Loshchilov and Frank Hutter. 2017.
\newblock Decoupled weight decay regularization.
\newblock \emph{arXiv preprint arXiv:1711.05101}.

\bibitem[{Lu et~al.(2023)Lu, Zhu, Han, Zhao, Mac~Namee, and Tan}]{lu-etal-2023-makes}
Jinghui Lu, Dongsheng Zhu, Weidong Han, Rui Zhao, Brian Mac~Namee, and Fei Tan. 2023.
\newblock \href {https://doi.org/10.18653/v1/2023.acl-long.128} {What makes pre-trained language models better zero-shot learners?}
\newblock In \emph{Proceedings of the 61st Annual Meeting of the Association for Computational Linguistics (Volume 1: Long Papers)}, pages 2288--2303, Toronto, Canada. Association for Computational Linguistics.

\bibitem[{Ma et~al.(2022)Ma, Tan, Zhou, Chen, Liang, Wang, Wu, Gui, and Zhang}]{ma2022searching}
Ruotian Ma, Yiding Tan, Xin Zhou, Xuanting Chen, Di~Liang, Sirui Wang, Wei Wu, Tao Gui, and Qi~Zhang. 2022.
\newblock Searching for optimal subword tokenization in cross-domain ner.
\newblock \emph{arXiv preprint arXiv:2206.03352}.

\bibitem[{McCloskey and Cohen(1989)}]{mccloskey1989catastrophic}
Michael McCloskey and Neal~J Cohen. 1989.
\newblock Catastrophic interference in connectionist networks: The sequential learning problem.
\newblock In \emph{Psychology of learning and motivation}, volume~24, pages 109--165. Elsevier.

\bibitem[{Mu and Lin(2025)}]{mu2025comprehensivesurveymixtureofexpertsalgorithms}
Siyuan Mu and Sen Lin. 2025.
\newblock \href {https://arxiv.org/abs/2503.07137} {A comprehensive survey of mixture-of-experts: Algorithms, theory, and applications}.
\newblock \emph{Preprint}, arXiv:2503.07137.

\bibitem[{Neyshabur et~al.(2015)Neyshabur, Tomioka, and Srebro}]{neyshabur2015search}
Behnam Neyshabur, Ryota Tomioka, and Nathan Srebro. 2015.
\newblock In search of the real inductive bias: On the role of implicit regularization in deep learning.
\newblock In \emph{Advances in Neural Information Processing Systems (NeurIPS)}.

\bibitem[{Ouyang et~al.(2022)Ouyang, Wu, Jiang, Almeida, Wainwright, Mishkin, Zhang, Agarwal, Slama, Ray et~al.}]{ouyang2022training}
Long Ouyang, Jeffrey Wu, Xu~Jiang, Diogo Almeida, Carroll Wainwright, Pamela Mishkin, Chong Zhang, Sandhini Agarwal, Katarina Slama, Alex Ray, and 1 others. 2022.
\newblock Training language models to follow instructions with human feedback.
\newblock \emph{Advances in neural information processing systems}, 35:27730--27744.

\bibitem[{Parmar et~al.(2024)Parmar, Satheesh, Patwary, Shoeybi, and Catanzaro}]{parmar2024reuse}
Jupinder Parmar, Sanjev Satheesh, Mostofa Patwary, Mohammad Shoeybi, and Bryan Catanzaro. 2024.
\newblock Reuse, don't retrain: A recipe for continued pretraining of language models.
\newblock \emph{arXiv preprint arXiv:2407.07263}.

\bibitem[{Peng et~al.(2023)Peng, Li, He, Galley, and Gao}]{peng2023instructiontuninggpt4}
Baolin Peng, Chunyuan Li, Pengcheng He, Michel Galley, and Jianfeng Gao. 2023.
\newblock \href {https://arxiv.org/abs/2304.03277} {Instruction tuning with gpt-4}.
\newblock \emph{Preprint}, arXiv:2304.03277.

\bibitem[{Pfeiffer et~al.(2020)Pfeiffer, Glaude, Gururangan, Lin, Kamath, Rücklé, Vulić, Gurevych, and Cho}]{pfeiffer2020adapterfusion}
Jonas Pfeiffer, Anaïs Glaude, Suchin Gururangan, Xi~Victoria Lin, Aishwarya Kamath, Andreas Rücklé, Ivan Vulić, Iryna Gurevych, and Kyunghyun Cho. 2020.
\newblock Adapterfusion: Non-destructive task composition for transfer learning.
\newblock In \emph{European Conference on Artificial Intelligence (ECAI)}.

\bibitem[{Qi et~al.(2024)Qi, Chen, Wang, Liu, Zheng, and Wang}]{qi2024optimizing}
Zhen Qi, Jiajing Chen, Shuo Wang, Bingying Liu, Hongye Zheng, and Chihang Wang. 2024.
\newblock Optimizing multi-task learning for enhanced performance in large language models.
\newblock \emph{arXiv preprint arXiv:2412.06249}.

\bibitem[{Raffel et~al.(2020)Raffel, Shazeer, Roberts, Lee, Narang, Matena, Zhou, Li, and Liu}]{raffel2020exploring}
Colin Raffel, Noam Shazeer, Adam Roberts, Katherine Lee, Sharan Narang, Michael Matena, Yanqi Zhou, Wei Li, and Peter~J Liu. 2020.
\newblock Exploring the limits of transfer learning with a unified text-to-text transformer.
\newblock \emph{Journal of machine learning research}, 21(140):1--67.

\bibitem[{Ren et~al.(2020)Ren, Guo, Lu, Zhou, Liu, Tang, Sundaresan, Zhou, Blanco, and Ma}]{ren2020codebleu}
Shuo Ren, Daya Guo, Shuai Lu, Long Zhou, Shujie Liu, Duyu Tang, Neel Sundaresan, Ming Zhou, Ambrosio Blanco, and Shuai Ma. 2020.
\newblock Codebleu: a method for automatic evaluation of code synthesis.
\newblock \emph{arXiv preprint arXiv:2009.10297}.

\bibitem[{Rolnick et~al.(2019)Rolnick, Ahuja, Schwarz, Lillicrap, and Wayne}]{rolnick2019experience}
David Rolnick, Aanuj Ahuja, Jonathan Schwarz, Timothy Lillicrap, and Gregory Wayne. 2019.
\newblock Experience replay for continual learning.
\newblock In \emph{Advances in Neural Information Processing Systems (NeurIPS)}.

\bibitem[{Sanh et~al.(2021)Sanh, Webson, Raffel, Bach, Sutawika, Alyafeai, Chaffin, Stiegler, Scao, Raja et~al.}]{sanh2021multitask}
Victor Sanh, Albert Webson, Colin Raffel, Stephen~H Bach, Lintang Sutawika, Zaid Alyafeai, Antoine Chaffin, Arnaud Stiegler, Teven~Le Scao, Arun Raja, and 1 others. 2021.
\newblock Multitask prompted training enables zero-shot task generalization.
\newblock \emph{arXiv preprint arXiv:2110.08207}.

\bibitem[{Solano et~al.(2024)Solano, Sanni, Camburu, and Minervini}]{solano2024sparsefitfewshotpromptingsparse}
Jesus Solano, Mardhiyah Sanni, Oana-Maria Camburu, and Pasquale Minervini. 2024.
\newblock \href {https://arxiv.org/abs/2305.13235} {Sparsefit: Few-shot prompting with sparse fine-tuning for jointly generating predictions and natural language explanations}.
\newblock \emph{Preprint}, arXiv:2305.13235.

\bibitem[{Song et~al.(2022)Song, Liang, Li, Li, Wang, Peng, Wu, and Yu}]{song2022improving}
Jian Song, Di~Liang, Rumei Li, Yuntao Li, Sirui Wang, Minlong Peng, Wei Wu, and Yongxin Yu. 2022.
\newblock Improving semantic matching through dependency-enhanced pre-trained model with adaptive fusion.
\newblock \emph{arXiv preprint arXiv:2210.08471}.

\bibitem[{Stickland and Murray(2020)}]{stickland2020bertpkd}
Ashton Stickland and Iain Murray. 2020.
\newblock Bert and pals: Projected attention layers for efficient adaptation in multi-task learning.
\newblock In \emph{International Conference on Machine Learning (ICML)}.

\bibitem[{Taori et~al.(2023)Taori, Gulrajani, Zhang, Dubois, Li, Guestrin, Liang, and Hashimoto}]{taori2023stanford}
Rohan Taori, Ishaan Gulrajani, Tianyi Zhang, Yann Dubois, Xuechen Li, Carlos Guestrin, Percy Liang, and Tatsunori~B Hashimoto. 2023.
\newblock Stanford alpaca: An instruction-following llama model.

\bibitem[{Team et~al.(2025)Team, Kamath, Ferret, Pathak, Vieillard, Merhej, Perrin, Matejovicova, Ramé, Rivière, Rouillard, Mesnard, Cideron, bastien Grill, Ramos, Yvinec, Casbon, Pot, Penchev, Liu, Visin, Kenealy, Beyer, Zhai, Tsitsulin, Busa-Fekete, Feng, Sachdeva, Coleman, Gao, Mustafa, Barr, Parisotto, Tian, Eyal, Cherry, Peter, Sinopalnikov, Bhupatiraju, Agarwal, Kazemi, Malkin, Fernandez, Newlan, yeong Ji, Singh, Black, Yu, Hui, Vodrahalli, Greff, Qiu, Valentine, Coelho, Ritter, Hoffman, Watson, Chaturvedi, Moynihan, Ma, Babar, Noy, Byrd, Roy, Momchev, Chauhan, Sachdeva, Zhang, Liu, Yacovone, Liechty, Kalra, Evci, Misra, Roseberry, Feinberg, Kolesnikov, Han, Kwon, Chen, Chow, Zhu, Wei, Egyed, Cotruta, Giang, Kirk, Rao, Black, Babar, Lo, Moreira, Martins, Sanseviero, Gonzalez, Gleicher, Warkentin, Mirrokni, Senter, Collins, Barral, Ghahramani, Hadsell, Matias, Sculley, Petrov, Fiedel, Shazeer, Vinyals, Dean, Hassabis, Kavukcuoglu, Farabet, Buchatskaya, Alayrac, Anil, Dmitry, Lepikhin, Borgeaud, Bachem,
  Joulin, Andreev, Hardin, Dadashi, and Hussenot}]{team2024gemma}
Gemma Team, Aishwarya Kamath, Johan Ferret, Shreya Pathak, Nino Vieillard, Ramona Merhej, Sarah Perrin, Tatiana Matejovicova, Alexandre Ramé, Morgane Rivière, Louis Rouillard, Thomas Mesnard, Geoffrey Cideron, Jean bastien Grill, Sabela Ramos, Edouard Yvinec, Michelle Casbon, Etienne Pot, Ivo Penchev, and 106 others. 2025.
\newblock \href {https://arxiv.org/abs/2503.19786} {Gemma 3 technical report}.
\newblock \emph{Preprint}, arXiv:2503.19786.

\bibitem[{Touvron et~al.(2023)Touvron, Lavril, Izacard, Martinet, Lachaux, Lacroix, Rozi{\`e}re, Goyal, Hambro, Azhar et~al.}]{touvron2023llama}
Hugo Touvron, Thibaut Lavril, Gautier Izacard, Xavier Martinet, Marie-Anne Lachaux, Timoth{\'e}e Lacroix, Baptiste Rozi{\`e}re, Naman Goyal, Eric Hambro, Faisal Azhar, and 1 others. 2023.
\newblock Llama: Open and efficient foundation language models.
\newblock \emph{arXiv preprint arXiv:2302.13971}.

\bibitem[{Wang et~al.(2022)Wang, Liang, Song, Li, and Wu}]{wang2022dabert}
Sirui Wang, Di~Liang, Jian Song, Yuntao Li, and Wei Wu. 2022.
\newblock Dabert: Dual attention enhanced bert for semantic matching.
\newblock \emph{arXiv preprint arXiv:2210.03454}.

\bibitem[{Wei et~al.(2021)Wei, Bosma, Zhao, Guu, Yu, Lester, Du, Dai, and Le}]{wei2021finetuned}
Jason Wei, Maarten Bosma, Vincent~Y Zhao, Kelvin Guu, Adams~Wei Yu, Brian Lester, Nan Du, Andrew~M Dai, and Quoc~V Le. 2021.
\newblock Finetuned language models are zero-shot learners.
\newblock \emph{arXiv preprint arXiv:2109.01652}.

\bibitem[{Wei et~al.(2022)Wei, Bosma, Zhao, Guu, Yu, Lester, Du, Dai, and Le}]{wei2022finetunedlanguagemodelszeroshot}
Jason Wei, Maarten Bosma, Vincent~Y. Zhao, Kelvin Guu, Adams~Wei Yu, Brian Lester, Nan Du, Andrew~M. Dai, and Quoc~V. Le. 2022.
\newblock \href {https://arxiv.org/abs/2109.01652} {Finetuned language models are zero-shot learners}.
\newblock \emph{Preprint}, arXiv:2109.01652.

\bibitem[{Wu et~al.(2025{\natexlab{a}})Wu, Qian, Liu, Wang, Huang, Liang, Miao, Dou, Lv, Wang et~al.}]{wu2025progressive}
Muling Wu, Qi~Qian, Wenhao Liu, Xiaohua Wang, Zisu Huang, Di~Liang, LI~Miao, Shihan Dou, Changze Lv, Zhenghua Wang, and 1 others. 2025{\natexlab{a}}.
\newblock Progressive mastery: Customized curriculum learning with guided prompting for mathematical reasoning.
\newblock \emph{arXiv preprint arXiv:2506.04065}.

\bibitem[{Wu et~al.(2024)Wu, Luo, Li, Pan, Vu, and Haffari}]{wu2024continuallearninglargelanguage}
Tongtong Wu, Linhao Luo, Yuan-Fang Li, Shirui Pan, Thuy-Trang Vu, and Gholamreza Haffari. 2024.
\newblock \href {https://arxiv.org/abs/2402.01364} {Continual learning for large language models: A survey}.
\newblock \emph{Preprint}, arXiv:2402.01364.

\bibitem[{Wu et~al.(2025{\natexlab{b}})Wu, Yang, Chai, Zhang, Liu, Du, Liang, Shu, Cheng, Sun et~al.}]{wu2025tablebench}
Xianjie Wu, Jian Yang, Linzheng Chai, Ge~Zhang, Jiaheng Liu, Xeron Du, Di~Liang, Daixin Shu, Xianfu Cheng, Tianzhen Sun, and 1 others. 2025{\natexlab{b}}.
\newblock Tablebench: A comprehensive and complex benchmark for table question answering.
\newblock In \emph{Proceedings of the AAAI Conference on Artificial Intelligence}, volume~39, pages 25497--25506.

\bibitem[{Wu et~al.(2025{\natexlab{c}})Wu, Yang, Li, Zhang, Du, Chai, Liang, and Li}]{wu2025unleashing}
Xianjie Wu, Jian Yang, Tongliang Li, Shiwei Zhang, Yiyang Du, LinZheng Chai, Di~Liang, and Zhoujun Li. 2025{\natexlab{c}}.
\newblock Unleashing potential of evidence in knowledge-intensive dialogue generation.
\newblock In \emph{ICASSP 2025-2025 IEEE International Conference on Acoustics, Speech and Signal Processing (ICASSP)}, pages 1--5. IEEE.

\bibitem[{Xu et~al.(2020)Xu, Zhang, Mao, Wang, Xie, and Zhang}]{xu-etal-2020-curriculum}
Benfeng Xu, Licheng Zhang, Zhendong Mao, Quan Wang, Hongtao Xie, and Yongdong Zhang. 2020.
\newblock \href {https://doi.org/10.18653/v1/2020.acl-main.542} {Curriculum learning for natural language understanding}.
\newblock In \emph{Proceedings of the 58th Annual Meeting of the Association for Computational Linguistics}, pages 6095--6104, Online. Association for Computational Linguistics.

\bibitem[{Xue et~al.(2024)Xue, Liang, Wang, and Zhang}]{xue2024question}
Chao Xue, Di~Liang, Pengfei Wang, and Jing Zhang. 2024.
\newblock Question calibration and multi-hop modeling for temporal question answering.
\newblock In \emph{Proceedings of the AAAI Conference on Artificial Intelligence}, volume~38, pages 19332--19340.

\bibitem[{Xue et~al.(2023)Xue, Liang, Wang, Zhang, and Wu}]{xue2023dual}
Chao Xue, Di~Liang, Sirui Wang, Jing Zhang, and Wei Wu. 2023.
\newblock Dual path modeling for semantic matching by perceiving subtle conflicts.
\newblock In \emph{ICASSP 2023-2023 IEEE International Conference on Acoustics, Speech and Signal Processing (ICASSP)}, pages 1--5. IEEE.

\bibitem[{Yang et~al.(2024)Yang, Yang, Hui, Zheng, Yu, Zhou, Li, Li, Liu, Huang, Dong, Wei, Lin, Tang, Wang, Yang, Tu, Zhang, Ma, Yang, Xu, Zhou, Bai, He, Lin, Dang, Lu, Chen, Yang, Li, Xue, Ni, Zhang, Wang, Peng, Men, Gao, Lin, Wang, Bai, Tan, Zhu, Li, Liu, Ge, Deng, Zhou, Ren, Zhang, Wei, Ren, Liu, Fan, Yao, Zhang, Wan, Chu, Liu, Cui, Zhang, Guo, and Fan}]{yang2024qwen2}
An~Yang, Baosong Yang, Binyuan Hui, Bo~Zheng, Bowen Yu, Chang Zhou, Chengpeng Li, Chengyuan Li, Dayiheng Liu, Fei Huang, Guanting Dong, Haoran Wei, Huan Lin, Jialong Tang, Jialin Wang, Jian Yang, Jianhong Tu, Jianwei Zhang, Jianxin Ma, and 43 others. 2024.
\newblock \href {https://arxiv.org/abs/2407.10671} {Qwen2 technical report}.
\newblock \emph{Preprint}, arXiv:2407.10671.

\bibitem[{Yang et~al.(2023)Yang, Pan, Wang, Yu, Shen, Chen, Xiao, Jiang, and Guo}]{Yang_2023}
Enneng Yang, Junwei Pan, Ximei Wang, Haibin Yu, Li~Shen, Xihua Chen, Lei Xiao, Jie Jiang, and Guibing Guo. 2023.
\newblock \href {https://doi.org/10.1609/aaai.v37i9.26275} {Adatask: A task-aware adaptive learning rate approach to multi-task learning}.
\newblock \emph{Proceedings of the AAAI Conference on Artificial Intelligence}, 37(9):10745–10753.

\bibitem[{Yu et~al.(2020)Yu, Kumar, Gupta, Levine, Hausman, and Finn}]{yu2020gradient}
Tianhe Yu, Saurabh Kumar, Abhishek Gupta, Sergey Levine, Karol Hausman, and Chelsea Finn. 2020.
\newblock Gradient surgery for multi-task learning.
\newblock \emph{Advances in neural information processing systems}, 33:5824--5836.

\bibitem[{Zamir et~al.(2018)Zamir, Sax, Shen, Guibas, Malik, and Savarese}]{zamir2018taskonomydisentanglingtasktransfer}
Amir Zamir, Alexander Sax, William Shen, Leonidas Guibas, Jitendra Malik, and Silvio Savarese. 2018.
\newblock \href {https://arxiv.org/abs/1804.08328} {Taskonomy: Disentangling task transfer learning}.
\newblock \emph{Preprint}, arXiv:1804.08328.

\bibitem[{Zhang et~al.(2021)Zhang, Norouzi, and Kolesnikov}]{zhang2021multitask}
Sheng Zhang, Mohammad Norouzi, and Alexander Kolesnikov. 2021.
\newblock Revisiting multi-task learning in the deep learning age.
\newblock \emph{arXiv preprint arXiv:2105.02178}.

\bibitem[{Zhang et~al.(2024)Zhang, Dong, Li, Zhang, Sun, Wang, Li, Hu, Zhang, Wu, and Wang}]{zhang2024instructiontuninglargelanguage}
Shengyu Zhang, Linfeng Dong, Xiaoya Li, Sen Zhang, Xiaofei Sun, Shuhe Wang, Jiwei Li, Runyi Hu, Tianwei Zhang, Fei Wu, and Guoyin Wang. 2024.
\newblock \href {https://arxiv.org/abs/2308.10792} {Instruction tuning for large language models: A survey}.
\newblock \emph{Preprint}, arXiv:2308.10792.

\bibitem[{Zhang et~al.(2025)Zhang, Wang, Hu, and Ma}]{zhang2025efficientknowledgetransfermultitask}
Xiao Zhang, Kangsheng Wang, Tianyu Hu, and Huimin Ma. 2025.
\newblock \href {https://arxiv.org/abs/2505.00009} {Efficient knowledge transfer in multi-task learning through task-adaptive low-rank representation}.
\newblock \emph{Preprint}, arXiv:2505.00009.

\bibitem[{Zheng et~al.(2023)Zheng, Chiang, Sheng, Zhuang, Wu, Lin, Li, Li, Xing et~al.}]{zheng2023judging}
Lianmin Zheng, Wei-Lin Chiang, Ying Sheng, Siyuan Zhuang, Zhanghao Wu, Zi~Lin, Zhuohan Li, Dacheng Li, Eric Xing, and 1 others. 2023.
\newblock Judging llm-as-a-judge with mt-bench and chatbot arena.
\newblock \emph{Advances in Neural Information Processing Systems}.

\bibitem[{Zheng et~al.(2022)Zheng, Bao, Zhou, Liang, Wang, Wu, Gui, Zhang, and Huang}]{zheng2022robust}
Rui Zheng, Rong Bao, Yuhao Zhou, Di~Liang, Sirui Wang, Wei Wu, Tao Gui, Qi~Zhang, and Xuanjing Huang. 2022.
\newblock Robust lottery tickets for pre-trained language models.
\newblock \emph{arXiv preprint arXiv:2211.03013}.

\end{thebibliography}

\appendix


\section{Experiments Setup}
\label{sec:experiments}

We conducted extensive experiments to evaluate the effectiveness of the proposed Dynamic Parameter Isolation (DPI) framework. The primary goals of our evaluation are to determine whether DPI outperforms standard supervised fine-tuning (SFT) baselines, including multi-task and multi-stage methods, across diverse and conflicting tasks; to assess DPI's ability to mitigate the "seesaw effect" and catastrophic forgetting; to examine the sensitivity of DPI to hyperparameters such as the core percentage ($p$) and similarity threshold ($\tau$); and to analyze the impact of its dynamic freezing mechanism.

\noindent{\textbf{Datasets:}}We evaluate DPI on a diverse suite of publicly available datasets that represent structured reasoning, code generation, and open-ended instruction-following tasks. For mathematical reasoning, we use GSM8K \cite{cobbe2021training}, which evaluates multi-step reasoning through accuracy. For code generation, we use CodeAlpaca \cite{chaudhary2023code}, where performance is measured using CodeBLEU \cite{ren2020codebleu}. For logical reasoning, we use LogiQA \cite{liu2020logiqa}, which assesses logical consistency through accuracy scores. For general instruction-following and conversational abilities, we use Alpaca \cite{taori2023stanford} and UltraChat \cite{ding2023enhancing}, both evaluated using GPT-4-based scoring on a 1-to-10 scale \cite{zheng2023judging}. These datasets include a mix of structured tasks (e.g., GSM8K, LogiQA, CodeAlpaca) and open-ended tasks (e.g., Alpaca, UltraChat) to introduce potential conflicts in parameter specialization.
Each task is evaluated using its standard metric: accuracy for GSM8K and LogiQA, CodeBLEU for CodeAlpaca, and GPT-4 scoring for Alpaca and UltraChat. To provide a unified comparison across tasks, we also report a macro-average score (\textbf{Avg. Norm. Score}) by normalizing individual task scores to a common 0-10 scale.

\noindent{\textbf{Baselines:}}We compare DPI against three SFT baselines. (1) Full Multi-task SFT, where the model is fine-tuned on a uniform mixture of all datasets without task grouping or parameter isolation. (2) Multi-Stage SFT (Random Grouping), where tasks are randomly partitioned into $K=3$ stages and fine-tuned sequentially, updating all parameters across each stage. (3) Multi-Stage SFT (Heuristic Grouping), where tasks grouped based on perceived similarity (e.g., reasoning tasks grouped together, open-ended tasks grouped together) are fine-tuned sequentially over two stages, with all parameters updated during each stage.

\noindent{\textbf{Implement details:}}All experiments use the LLaMA-2-7B \cite{touvron2023llama}, Mistral-8B \cite{jiang2024mixtral}, Gemma-9B \cite{team2024gemma}, and Qwen2-7B \cite{yang2024qwen2} as the base language model. Fine-tuning is performed using the AdamW optimizer \cite{loshchilov2017decoupled} with a learning rate of $1 \times 10^{-5}$, batch size of 64, and a cosine learning rate schedule with 3\% warm-up steps. The main SFT stages involve training for three epochs on the datasets for each stage. For DPI, core parameter identification is conducted through probe fine-tuning runs lasting one epoch per task ($E_{\text{probe}}=1$). DPI hyperparameters are set to a core percentage of $p=1\%$ and a similarity threshold $\tau=0.1$. Task groups derived by DPI are randomly ordered for multi-stage training. Masked fine-tuning is applied during each stage, leveraging the dynamic freezing mechanism to preserve task-specific core parameter regions. All experiments are performed on machines equipped with eight NVIDIA A100 GPUs (80GB).

\section{Related Work}
\label{sec:related}

\subsection{Supervised Fine-Tuning and Instruction Tuning}
Supervised Fine-Tuning (SFT) is a prevalent technique for specializing pre-trained LLMs \cite{brown2020language,devlin2019bert, raffel2020exploring} for desired downstream behaviors. Instruction tuning \cite{wei2021finetuned, sanh2021multitask, chung2024scaling, zheng2022robust, ma2022searching, wu2025progressive,ouyang2022training,zhang2024instructiontuninglargelanguage,anthropic2024,peng2023instructiontuninggpt4,fei2022cqg,liu2023time,liu2023local,xue2024question} a specific form of SFT, leverages datasets formatted as instructions and responses to enhance model controllability and generalization to unseen tasks. Standard SFT often involves training on a mixture of data from various tasks \cite{longpre2023flan,chung2022scalinginstructionfinetunedlanguagemodels,wei2022finetunedlanguagemodelszeroshot,lu-etal-2023-makes,wu2025tablebench,wu2025unleashing}, typically applying updates across the entire parameter space. While effective for general adaptation, this approach can struggle when task objectives within the SFT data mixture conflict, leading to the performance trade-offs ("seesaw effect") discussed in Section~\ref{sec:introduction}. Our work diverges from this standard practice by proposing a method to selectively update parameters based on their identified relevance to specific tasks within the SFT process, thereby directly addressing the negative consequences of indiscriminate parameter updates.

\subsection{Task Interference and Knowledge Retention}

Task interference has been a persistent challenge in multi-task and sequential learning paradigms. Specifically, training sequentially on multiple tasks often leads to catastrophic forgetting \cite{mccloskey1989catastrophic,wu2024continuallearninglargelanguage}, where knowledge acquired in earlier stages is overwritten or degraded by subsequent updates. This problem has been studied extensively in the context of smaller neural architectures and motivated approaches such as regularization \cite{kirkpatrick2017overcoming,zhang2025efficientknowledgetransfermultitask}, replay-based methods \cite{rolnick2019experience,liu2024multi}, and episodic memory mechanisms \cite{lopez2017gem,guo2018dynamic,zamir2018taskonomydisentanglingtasktransfer}. While effective for small-scale models, extending these techniques to the massive parameter spaces of LLMs is non-trivial.
Gradient-based methods have gained popularity for balancing task objectives. GradNorm \cite{chen2018gradnorm,lopezpaz2022gradientepisodicmemorycontinual} adjusts task-specific gradient magnitudes, while PCGrad \cite{yu2020gradient,Yang_2023} selectively projects conflicting gradients to mitigate interference during multi-task learning. These approaches focus primarily on optimizing task gradients without addressing the root cause of parameter-level contention. Modular solutions, such as adapter fusion \cite{pfeiffer2020adapterfusion,fedus2022switchtransformersscalingtrillion,xu-etal-2020-curriculum,li2017learningforgetting,liang2019adaptive,wang2022dabert,liang2019asynchronous}, assign independent modules to tasks, allowing architectural separation. Despite their advantages, such methods introduce added complexity and may not scale gracefully to hundreds  of tasks.
Our work departs from these paradigms by taking a parameter-centric approach. Rather than manipulating gradients or enforcing modularity, DPI directly quantifies and isolates the ``core parameters'' for each task based on update magnitudes. 

\begin{figure} 
    \centering
    \includegraphics[width=0.48\textwidth]{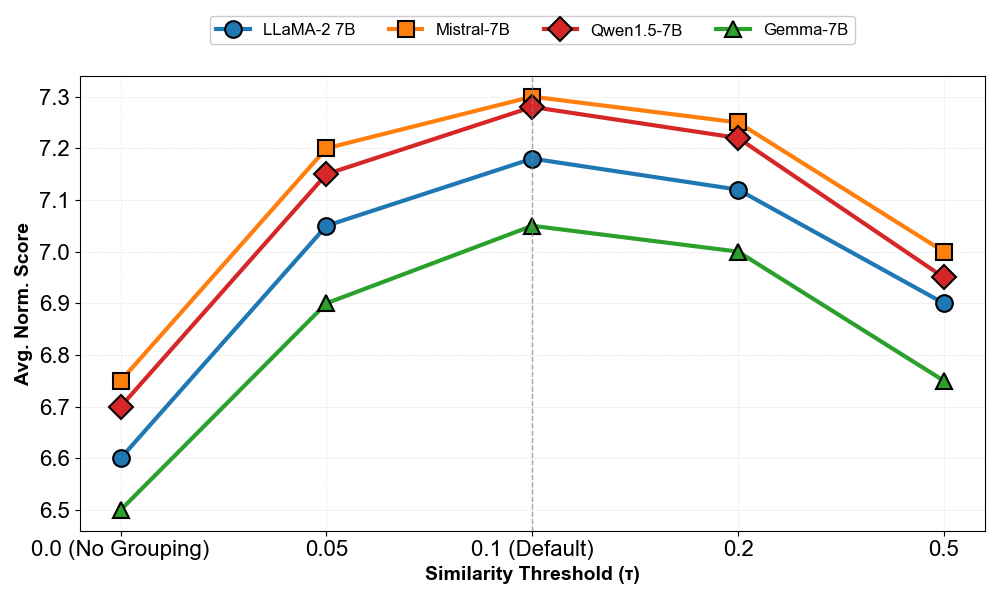}
    \caption{Ablation study on the similarity threshold $\tau$ across different base models (with $p=5\%$). Scores are Avg. Norm. Score. }
    \label{fig:figure_t}
\end{figure}

\subsection{Multi-Stage Fine-Tuning and Dynamic Scheduling}

Multi-stage fine-tuning frameworks have been widely adopted to tackle task heterogeneity in supervised tuning scenarios. These approaches often heuristically group tasks into stages based on shared characteristics, such as similarity or difficulty, and train sequentially across multiple phases \cite{ouyang2022training, wei2021finetuned}. While multi-stage frameworks can mitigate direct gradient conflict by separating tasks temporally, they do not account for overlaps in shared parameter usage. As a result, tasks in later stages can destructively overwrite knowledge embedded in parameters critical to earlier tasks, exacerbating catastrophic forgetting.
Multi-task fine-tuning strategies emphasize concurrent training on several tasks to enable shared representations \cite{caruana1997multitask}. However, multi-task learning encounters challenges in balancing competing task gradients due to loss imbalance or gradient directionality. Techniques combining shared and task-specific spaces, such as \cite{stickland2020bertpkd} and \cite{zhang2021multitask}, attempt to allocate independent regions of the model to different tasks. These approaches maintain task separation but often constrain model capacity and reduce the ability to leverage shared knowledge across tasks effectively.
Dynamic task scheduling has emerged as a promising approach to improve multi-task and multi-stage fine-tuning. For instance, \cite{chen2020just} proposed task prioritization based on difficulty, enabling the model to iteratively refine its understanding across task sequences. Similarly, \cite{aribandi2021ext5} presented heuristic strategies for task grouping and ordering to reduce task conflict. Although these methods improve task alignment through better scheduling, they still treat task interactions primarily at a coarse data level and do not address task-specific parameter differentiation within the LLMs.
Our approach, Dynamic Parameter Isolation (DPI), extends beyond these advancements by performing explicit parameter-level disentanglement. Unlike static task scheduling, our method dynamically identifies and freezes task-specific parameter regions during multi-stage fine-tuning. Tasks with overlapping parameter regions are grouped into joint training stages to maximize synergy, while disjoint tasks are staged sequentially with frozen core parameters from earlier stages. By aligning task scheduling explicitly with parameter sensitivity, DPI substantially mitigates destructive interference without the need for heuristic task grouping or modular constraints.

\subsection{Parameter Heterogeneity and Isolation}

The notion of parameter heterogeneity, where different parameters within a model contribute disproportionately to learning specific tasks, has been explored in the contexts of orthogonal weights, sparse updates, and parameter sharing \cite{neyshabur2015search, frankle2019winning,mu2025comprehensivesurveymixtureofexpertsalgorithms}. Inspired by these findings, parameter-efficient fine-tuning methods like adapters \cite{houlsby2019bertadapter,frantar2023sparsegptmassivelanguagemodels} and LoRA \cite{hu2021lora,solano2024sparsefitfewshotpromptingsparse,song2022improving,liu2024resolving,xue2023dual,li2024local, li2024comateformer} leverage this heterogeneity by introducing task-dedicated parameter subspaces. Such methods demonstrate that task-specific isolation can effectively reduce interference, but they generally require additional parameters, limiting scalability in resource-constrained applications.
A related line of research investigates parameter reuse and specialization within transfer learning and continual learning. \cite{parmar2024reuse,li-liang-2021-prefix} explored weight specialization during pretraining and task adaptation but did not explicitly quantify task-specific parameter regions. Conceptually closer to our work, \cite{qi2024optimizing} introduced task-sensitive routing to partition parameter updates among tasks dynamically. While this approach focuses on modular task routing, our framework operates directly on model parameter sensitivity and exploits organic updates of pre-trained LLMs.
DPI introduces a principled, data-driven approach to identifying and preserving task-specific core parameter regions within the same model architecture. 

\end{document}